\documentclass{ecai} 
\usepackage{latexsym}
\usepackage{amssymb}
\usepackage{amsmath}
\usepackage{amsthm}
\usepackage{subcaption}
\usepackage{booktabs}
\usepackage{enumitem}
\usepackage{graphicx}
\usepackage{hyperref}
\usepackage{color}

\newcommand{\BibTeX}{B\kern-.05em{\sc i\kern-.025em b}\kern-.08em\TeX}

\newcommand{\set}[1]{\ensuremath\mathcal{#1}}
\newcommand{\vect}[1]{\ensuremath\mathbf{#1}}
\newcommand{\card}{\ensuremath\vect{c}}
\newcommand{\mycards}{\ensuremath\set{C}}
\newcommand{\pack}{\ensuremath\set{P}}
\newcommand{\pref}{\succ}

\begin{document}

%%%%%%%%%%%%%%%%%%%%%%%%%%%%%%%%%%%%%%%%%%%%%%%%%%%%%%%%%%%%%%%%%%%%%%%%

\begin{frontmatter}

%%% Use this command to specify your submission number.
%%% In doubleblind mode, it will be printed on the first page.

\paperid{123} 

%%% Use this command to specify the title of your paper.

\title{Contrastive Learning of Preferences \\ 
%for Card Drafting 
with a Contextual InfoNCE Loss }

%%% Use this combinations of commands to specify all authors of your 
%%% paper. Use \fnms{} and \snm{} to indicate everyone's first names 
%%% and surname. This will help the publisher with indexing the 
%%% proceedings. Please use a reasonable approximation in case your 
%%% name does not neatly split into "first names" and "surname".
%%% Specifying your ORCID digital identifier is optional. 
%%% Use the \thanks{} command to indicate one or more corresponding 
%%% authors and their email address(es). If so desired, you can specify
%%% author contributions using the \footnote{} command.

\author[A]{\fnms{Timo}~\snm{Bertram}\thanks{Corresponding Author. Email: tbertram@faw.jku.at}}
\author[A]{\fnms{Johannes}~\snm{Fürnkranz}}
\author[B]{\fnms{Martin}~\snm{Müller}} 

\address[A]{Johannes Kepler University, Linz, Austria}
\address[B]{University of Alberta, Edmonton, Canada}

%%% Use this environment to include an abstract of your paper.

\begin{abstract}
A common problem in contextual preference ranking is that a single preferred action is compared against several choices, thereby blowing up the complexity and skewing the preference distribution.
In this work, we show how one can solve this problem via a suitable adaptation of the CLIP framework.
% to only allow restricted preferences while still resulting in strong empirical performance.
This adaptation is not entirely straight-forward, because although the InfoNCE loss used by CLIP has achieved great success in computer vision and multi-modal domains, its batch-construction technique requires the ability to compare arbitrary items, and is not well-defined if one item has multiple positive associations in the same batch. We empirically demonstrate the utility of our adapted version of the InfoNCE loss in the domain of collectable card games, where we aim to learn an embedding space that captures the associations between single cards and whole card pools based on human selections. Such selection data only exists for restricted choices, thus generating concrete preferences of one item over a set of other items rather than a perfect fit between the card and the pool.

Our results show that vanilla CLIP does not perform well due to the aforementioned intuitive issues. However, by adapting CLIP to the problem, we receive a model outperforming previous work trained with the triplet loss, while also alleviating problems associated with mining triplets.
\end{abstract}

\end{frontmatter}

\section{Introduction}

Preference Learning \citep{plbook} is a machine learning framework, where the task is to learn a function over a set of objects $\mathcal{O}$ which indicates a degree of preference for a given object $\vect{o}$. One way of solving this problem is to learn a numeric \emph{utility function} $u:\mathcal{O} \rightarrow \mathbb{R}$, where higher values of $u$ indicate a higher degree of preference, i.e., 
\begin{equation}
    u(\vect{o}_1) > u(\vect{o}_2) \Leftrightarrow \vect{o}_1 \pref \vect{o}_2
\end{equation}
In many cases, such a function is learned from a training set $\mathcal{D}$ of binary constraints of the form $(\vect{p}_i \pref \vect{n}_i), i = 1 \dots |\mathcal{D}|$, where the $i$-th constraint indicates that object $\vect{p}_i$ is preferred over $\vect{n}_i$. In the following, we will often call $\vect{p}_i$ the \emph{positive} and $\vect{n}_i$ the negative example of the $i$-th preference.

As there are $O(|\mathcal{O}|^2)$ possible preference constraints over the set of objects $\mathcal{O}$, it is difficult to manually label a sufficient number of training examples. However, in many cases, preferences can be observed from behavioural traces. In particular, in decision-making processes, one can infer preferences from observations, in which one action has been selected (and thus preferred) over all alternative actions. Attempts for formalizing such settings in the context of Markov decision processes can be found in the literature \citep{Ordinal-MDPs,cw:JMLR-17}.

Moreover, in many cases, the observed preference depends on a given context. A classic example is that the wine preference may depend on the choice of the accompanying main dinner course. Reasoning with such \emph{contextual preferences} have, e.g., been formalized in \emph{CP-nets} \citep{CP-nets}.
A simple example for such a setting is \emph{label ranking} \citep{jf:AIJ,plbook:Vembu}, where contexts are given in the form of training examples, for which preference constraints are defined over (some of) the possible labels.
However, in general, the objects over which preferences are defined may also be more complex than simple labels. For example, in the application we are working on, we intend to learn preferences over playing cards in the context of the cards a player is already holding. This allows a player to select from an available, restricted set of cards, where the best selection is not necessarily the overall best card, but the one that is the best addition to the set of cards that the player is already holding---a property that is crucial for success in collectable card games. Similar problems exist in many domains, such as, e.g., team building (the best addition to a team may not necessarily be the best player available on the market), recommender systems (the best recommendation for a user will not be the most popular book but the book that best fits the set of books they have already bought), etc.

A common problem in learning such a contextual preference function from behavioural traces is that one typically observes one preferred out of a larger set of possible actions. In our case, we can observe the card that a user has picked out of a set of dozens of cards they could have selected. This gives rise to a set of pairwise preferences where a single picked card is paired with $r$ rejected cards. Not only could $r$ be a very large number, but it may also result in a skewed sample distribution because the single positive example will appear with higher frequency than the $r$ negative examples.

In this paper, we improve upon prior work \citep{bertram2021predicting,tb:CoG-24} that used contextual preference ranking in the context of the collectable card game \emph{Magic: The Gathering}. Our key contribution is to demonstrate how the above-mentioned setting can be solved by converting $r$ pairwise contextual preferences into a single comparison of over $r$ objects. To this end, we adapt the well-known CLIP framework \citep{radford2021learning} to this task, showing that while a straight-forward adaption does not work as intended, a few minor modifications allow us to not only improve the predictive performance but also speed up the training process, while not needing to mine triplets online.

%\section{Related Work}
% Not sure we need this...

\begin{figure*}[ht]
    \centering
    \begin{subfigure}[t]{0.48\textwidth}
        \centering
        \includegraphics[width = \textwidth]{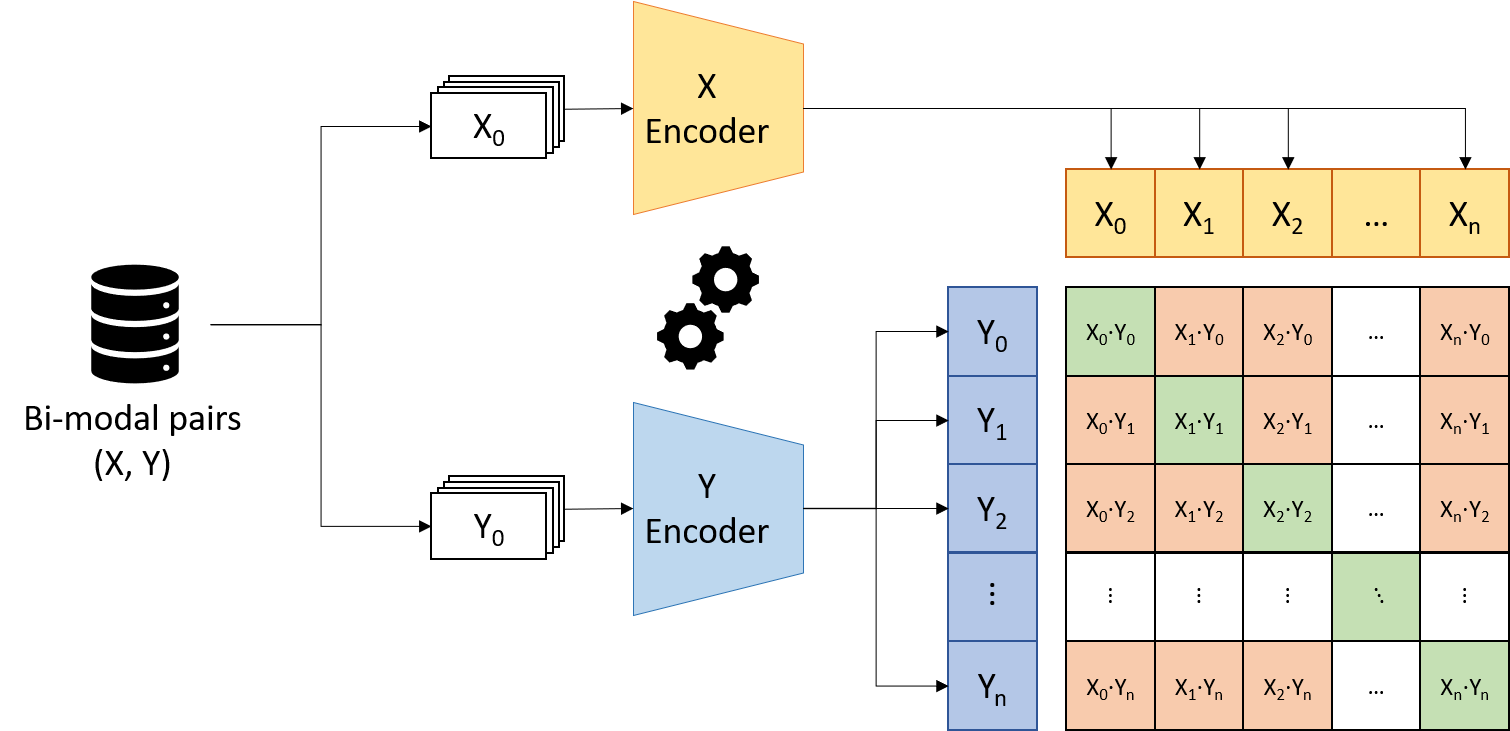}
    \caption{Contrastive learning approach with InfoNCE loss as in CLIP \cite{radford2021learning}. Positive examples are aligned on the diagonal, all other combinations serve as negative examples.}
    \end{subfigure}%
    \hspace{4mm}
    \begin{subfigure}[t]{0.48\textwidth}
        \centering
        \includegraphics[width=\textwidth]{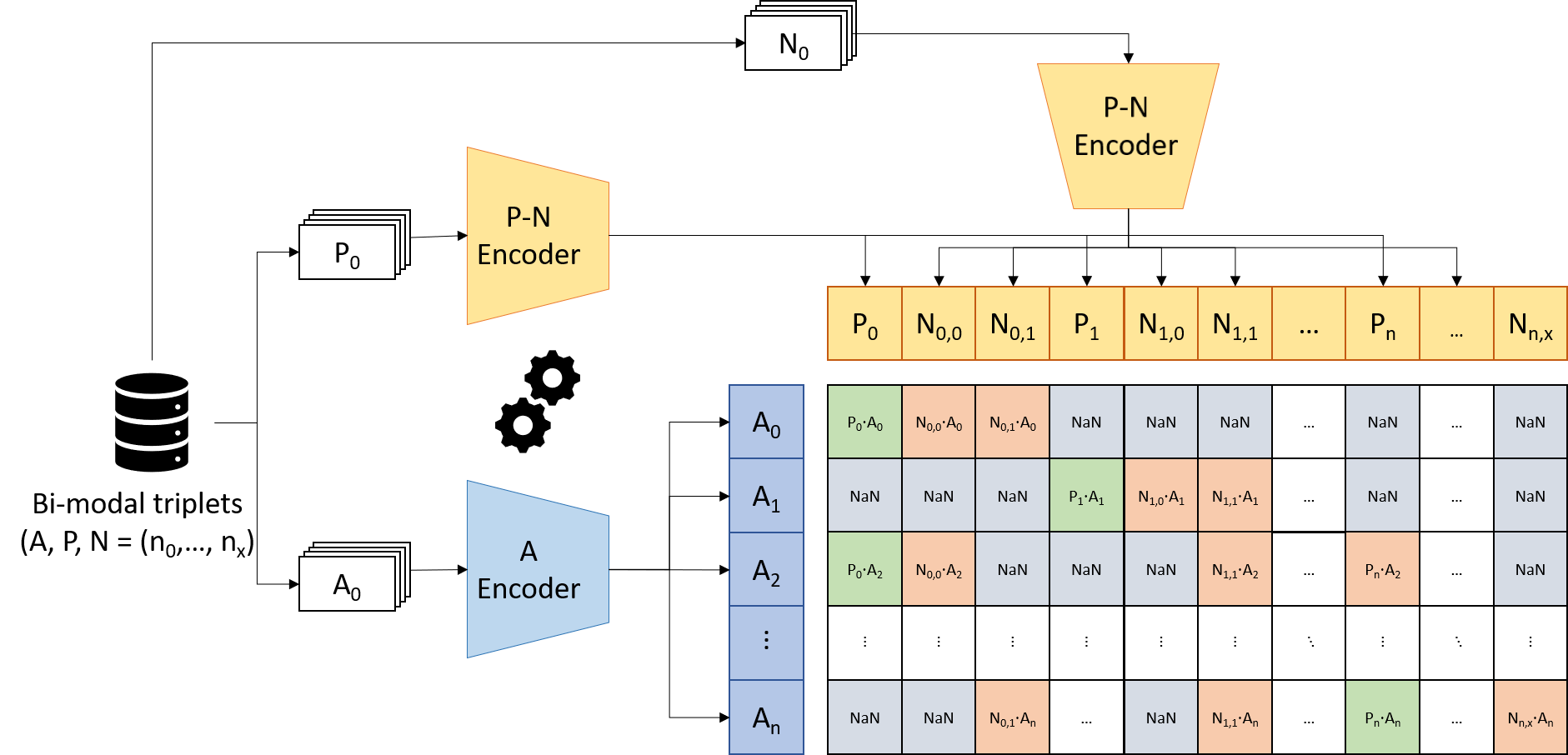}
    \caption{Proposed contrastive learning approach with masked InfoNCE loss. In practice, the whole similarity matrix is computed and a second masking matrix is generated, we omit this for brevity. The final matrix only contains the explicit comparisons from triplets.}
    \end{subfigure}
    \vspace{4mm}
    \caption{High-level overview of the changes necessary to accommodate contextual comparisons rather than arbitrary ones. }
    \label{fig:overview}
\end{figure*}

We start with a brief introduction to our 
application setting, card selection in a collectable card game, in Section~\ref{sec:Magic}. Section~\ref{sec:Contribution} introduces our main contribution, an adaption of CLIP to the setting of contextual preference ranking, and the results of an empirical evaluation of this approach are shown in Section~\ref{sec:Experiments}.

\section{Deckbuilding in Collectable Card Games}
\label{sec:Magic}

Collectable card games are a significant portion of the board game market and offer a large competitive scene. Games such as \textit{Hearthstone} or \textit{Magic: The Gathering} possess player counts in the millions and \textit{Magic: The Gathering} is currently evaluated at over \$1 billion\footnote{https://investor.hasbro.com/magic-gathering}. Still, game AI for those games is still in its infancy and outside of vastly simplified game versions \cite{xi2023mastering}, few models exist that are able to navigate aspects of them. In this work, we regard one subarea of constructing sophisticated collectable card game agents; \textbf{deckbuilding}.

\subsection{Deckbuilding}

\textbf{Deckbuilding}, i.e. selecting which cards to play the game with, is an integral aspect of every collectable card game. The specific choice of cards crucially enables gameplay and meaningful selections are necessary to compete against others. Still, they are also influenced by personal preference and card availability, allowing for much discussion and uncertainty. So far, constructing well-versed models to build functional decks on a similar level as humans is still a work in progress \cite{xiao2023mastering}. 

In most game versions, building a deck is separate from playing with it, i.e. the deckbuilding process is not built into gameplay but rather occurs in advance. However, some game versions strongly restrict deckbuilding by integrating it into the gameplay and reducing the pool of cards to choose from. Such game modes are called \textit{Limited} games. We outline the process in detail in Section~\ref{sec:Limited}

\subsection{Limited deckbuilding}
\label{sec:Limited}

In \textbf{Limited}, players sequentially build their decks based on a limited selection of cards. This process is called \textbf{drafting}.

\begin{itemize}
    \item For every decision point, a player is offered a set of cards called a \textbf{pack}. At the start of the process, it consists of 15 unique cards.
    \item The player selects a single card and adds it to their \textbf{pool}. After selecting the card, the player passes all unchosen cards to a neighbouring opponent. Note that \textbf{pool} and \textbf{deck} are interchangeable terms for the same concept; the set of cards a player has already chosen.
    \item The player receives a separate pack from the opponent on the other side. As that opponent has previously picked a card, this pack contains one less card. The player again selects a single card and passes the remaining.
    \item After 15 selections, each with a decreasing number of options, all steps above are repeated twice with two additional packs.
\end{itemize}

Altogether, a player makes 42 selections, as for the last card in each of the three packs, no decision occurs. Disregarding the influence of one's own selections on the opponents and disregarding card duplicates, this leads to $3\cdot 15! > 10^{12}$ separate decks a player can end up with. While this might not seem like a large number, packs contain a semi-random distribution of roughly 250 unique cards. With this, the quantity of unique drafts is enormous. To add to the difficulty of learning this process, evaluating the chosen cards depends on the actual gameplay with them, which is a large and unsolved problem in itself. Thus, currently no strong drafting or playing against for full games of \textit{Magic: The Gathering} exist. In this work, we aim to work towards being able to generate drafting agents for this game by leveraging human data. Such agents could be used as opponents or training tools for less experienced players.

\subsection{Data}
\label{sec:data}
For this work, we use a dataset collected by \href{https://17lands.com}{17lands}. 17lands is a website focused on \textit{Limited} gameplay of \textit{Magic: The Gathering} and collects data on a large amount of experienced human players. The dataset we use consists of human selections of cards throughout the drafting process. For every draft, we receive information on the possible card choices ($\pack$) and which card was chosen ($\card^*$). As card choices are highly situational and strongly depend on previous decisions, we relate the given card choice to all cards a player has previously chosen $\mycards$. The following section explains how we train models with this data.

\clearpage
\section{Contextual Preference Ranking}
\label{sec:Contribution}

In this section, we introduce the main contribution of this paper. We first review the previous use of pairwise constraints for training a Siamese neural network for contextual preference ranking (Section~\ref{sec:Siamese}), then briefly recapitulate CLIP and the InfoNCE loss (Section~\ref{sec:CLIP}), and finally demonstrate how it can be adapted to our contextual ranking setting (Section~\ref{sec:Contextual_InfoNCE}).

\subsection{Siamese Neural Networks for Drafting}
\label{sec:Siamese}

As shown by \citet{bertram2021predicting}, Siamese Neural Networks (SNNs) \cite{bromley1993signature} can effectively be used to predict human player's decisions. In their \textbf{contextual preference ranking} (CPR) framework, several options are compared against each other in the context of a specific item.

Concretely, CPR learns to predict a preference of one choice $\card_j$ over $\card_k$ in the context of $\mycards$, i.e. 
\begin{equation}
\label{eq:preference}
        \left(\card_j \pref \card_k \mid \mycards\right).
\end{equation}

In practice, this is achieved by training an SNN with the triplet loss (Equation \ref{eq:triplet}) to generate a latent embedding space of $\card_j$, $\card_k$, and $\mycards$, in which $d(\card_j, \mycards) < d(\card_k, \mycards)$ according to some distance metric $d$ (e.g. the Euclidian distance).

\begin{equation}
\label{eq:triplet}
    L_{\textrm{triplet}}(\vect{a},\vect{p},\vect{n}) = \max\left(d(\vect{a},\vect{p})-d(\vect{a},\vect{n}) + m,0\right)
\end{equation}

Crucially, while the network is solely trained on comparisons of single choices, distances in the embedding space are transitive and we can thus compare an arbitrary amount of items, i.e.

\begin{equation}
\label{eq:transitive}
(\card_j \pref \card_k \mid \mycards)
\land
(\card_k \pref \card_l \mid \mycards)
\Rightarrow
(\card_j \pref \card_l \mid \mycards)
\end{equation}

This enables ranking a whole set of items in a given context, thus allowing us to use the method for drafting (Section~\ref{sec:Limited}). While this is highly related to using SNNs for image recognition \cite{schroff2015facenet}, training the SNN is more restricted here.

Given a dataset of images $D$ and a labelling function $f: \mathbb{R}^n \mapsto [1,C]$ one can construct numerous valid triplets $(a,p,n)$ by sampling random tuples $(x_0, x_1, x_2) $ from $D$ where $f(x_0) = f(x_1) \And f(x_0) \neq f(x_2) \And x_0 \neq x_1$. While not allowing for completely arbitrary comparisons, e.g. if $f(x_0) \neq f(x_1) \neq f(x_2)$ one can construct no triplet, a well-constructed dataset will provide plenty of possible comparisons even when mining batches online, as $a$, $p$, and $n$ are all simple images.

For our domain, only restricted comparisons apply. For given set of items $P$, the dataset is constructed by relating the human selection $\card^* \in \pack \mid \mycards$ to all other $\card_0,...,\card_n \in \pack$. As opposed to the previous domain, all $\card_i \not\in \pack$ can not be compared and thus do not provide valid triplets. Thus, for our work, we sample decisions of the form $(\mycards, \pack)$ with the labelling function $f: \pack \times \mycards \mapsto [1,15]$, from which we can generate the triplets ($\mycards, \pack[f(\pack, \mycards)],\card_i)$ for all $ i = 1, ..., |\pack|$. Note that $f$ is not necessarily deterministic, i.e. separate records in the dataset can choose different cards in the exact same situation.

\subsection{InfoNCE Loss in CLIP}
\label{sec:CLIP}
The recently popularised InfoNCE loss \cite{oord2018InfoNCE, radford2021learning} is a powerful tool for contrastive learning. This representation learning technique is based on matching pairs of different modalities, e.g. pairs of images with associated captions in CLIP.
For a batch of pairs, CLIP \cite{radford2021learning} aligns all possible combinations in a square matrix with the correct pairings on the diagonal, training the text-encoder and image-encoder to maximise the cosine similarity of the diagonal and minimising all others (see Figure~\ref{fig:overview}). However, this only results in perfect pairings when assuming that no other items in the batch interfere with the pairs by being more similar.
Given a batch of $N$ pairs of images $I$ with texts $T$, it is assumed that, for a given similarity measure $sim$:

\begin{equation}
    \forall i \in [0,N] \not \exists j \in [0,N]: sim(I_i,T_i) > sim(I_i, T_j), i \neq j
\end{equation}

However, for sufficiently large batches, it is not unlikely that multiple images fit a text or vice versa. In practice, this inaccuracy is likely mitigated by the large dataset, but using the same approach for smaller datasets or ones with frequently reoccurring items might lead to performance degradation. Our domain is one such example where there are only roughly 250 different cards, which frequently results in one batch containing the same items. Therefore, it is not possible to simply minimise the cross entropy loss in both dimensions of the pair-matrix, as there will be correct pairings not on the diagonal, which means that one card can be associated with multiple decks. Some later adaptions of CLIP \cite{zhai2023sigmoid} are able to handle such settings by circumventing the softmax normalisation and cross-entropy computation by using a pairwise sigmoid loss, but we show in Section~\ref{sec:Experiments} that our domain possesses additional difficulties which do not allow this simple change.

\subsection{Contextual InfoNCE for Drafting}
\label{sec:Contextual_InfoNCE}

\begin{figure}[h]
    \centering
    \includegraphics[width = \columnwidth]{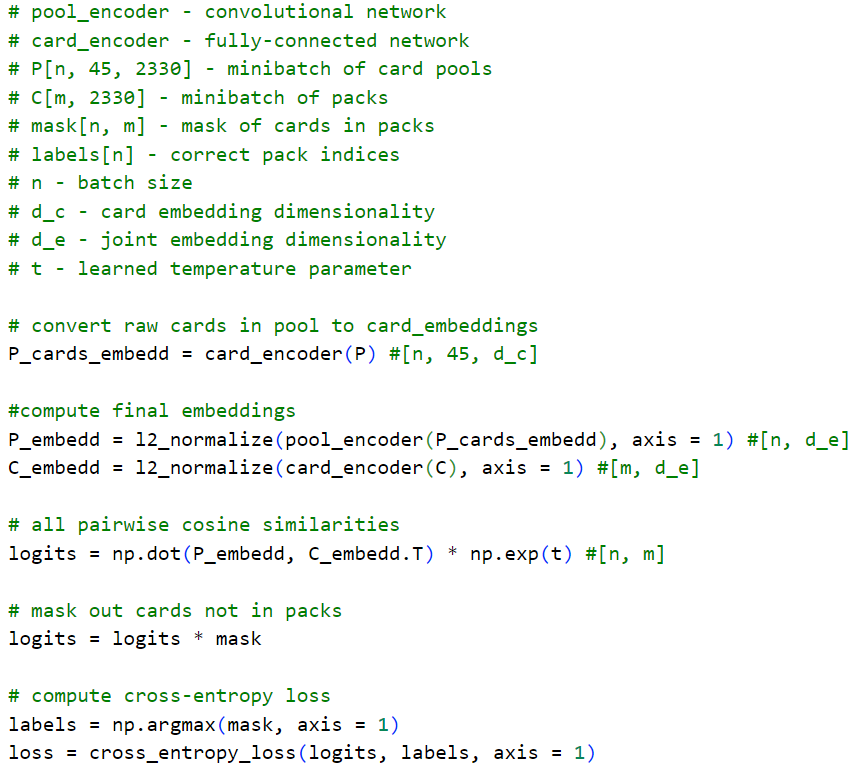}
    \caption{Pseudocode for the adapted InfoNCE loss, heavily inspired by the code in CLIP \cite{radford2021learning}.}
    \label{fig:pseudocode}
\end{figure}
\vspace{5mm}

In order to use the InfoNCE loss for drafting we need to make several adaptions. When training our model, we sample decisions from the dataset of the form $(\mycards, \pack)$ and have access to the labelling function $f$, where $f(\pack, \mycards) = \card^*$, i.e. we can identify the chosen card in the pack. We create a multi-modal embedding space which relates card pools $\mycards$ to cards $\card \in \pack$ by training a pool encoding network and a card encoding network. In contrast to the embeddings created in many other domains, where the space models the similarity of items, ours represents player decisions. Thus, we still aim to maximise the cosine similarity of the pool and chosen cards while minimising the cosine similarity of the pools and unchosen cards. 

However, the domain restrictions require us to construct batches differently (Figure~\ref{fig:overview}). CLIP samples $N$ labelled images and creates an $N \times N$ matrix of all possible combinations with the diagonal providing the correct pairs, which implicitly creates $N^2 -N$ incorrect pairs that serve as negative examples. Note that negative examples are not explicitly sampled. Rather, for one positive pair all combinations with other items in the matrix serve as negatives. For our task, this is not possible as we do not have information about the majority of combinations in such a matrix of strictly positive pairs. In addition, positive examples are based on human decisions and are thus strongly biased towards better cards, while weaker cards, for which it is important to learn their weakness, would rarely occur. 

Thus, instead of aligning pairs, we use all triplets in $(\mycards, \pack)$, including all negative examples. We create a $N \times M$ matrix where $N$ is the number of samples and $M$ is the number of unique cards in the dataset. In addition, we create a $N \times M$ mask which marks valid and invalid comparisons which is crucial to computing the loss. Each sampled decision provides one row of the matrix, where the card pool is related to the possible choices, with the mask having a 1 at the chosen card, 0s at unchosen cards and -1s at cards not in the offering. 

To summarise, our approach (see Figure~\ref{fig:overview}) learns an embedding space by maximising the similarity of the chosen card and the pool, while minimising the similarity of the pool and the unchosen cards. Cards not available in the pack are not used for the computation of the cross entropy loss and thus their similarity is irrelevant but with the batch construction into an $N \times M$ matrix, we can parallelise the computation. In contrast to CLIP, we do not compute the column-wise cross-entropy as this disregards the available cards and is thus not applicable to the domain.

\section{Experiments}
\label{sec:Experiments}

In this section, we compare our proposed method to previous research using the triplet loss and an unaltered version of CLIP. For the triplet approach, we use the research \citet{bertram2021predicting}, CLIP-based experiments were re-implemented by us. Although the specific implementation details are not of high importance, as we aim to equalise hyperparameters, we briefly cover implementation details in Section~\ref{sec:architecture}. For all experiments, we use the datasets provided by \href{https://17lands.com}{17 lands}. These datasets provide turn-by-turn records of human card selections when drafting. We regard these drafting decisions are preferences of one card over a set of other cards given that player's pool. Thus, we can directly generate training samples, i.e. triplets of the pool, chosen card and unchosen cards, from the dataset. While numerous datasets with different cards are available, we arbitrarily chose a single one (\href{https://www.17lands.com/public_datasets}{NEO}). This dataset consists of 5,693,460 samples which we split 80/20 into training and test data. 

All outlined training approaches aim to model the contextual card preferences by learning an embedding space in which preferred cards possess embeddings with high cosine similarity, or small distance, to the pool. The methods mainly differ by the loss function used and how the training data is processed by the networks.

\subsection{Architectures and Hyperparameters}
\label{sec:architecture}

\begin{figure*}[ht]
    \centering
    \begin{subfigure}[t]{0.5\textwidth}
        \centering
        \includegraphics[width = \textwidth]{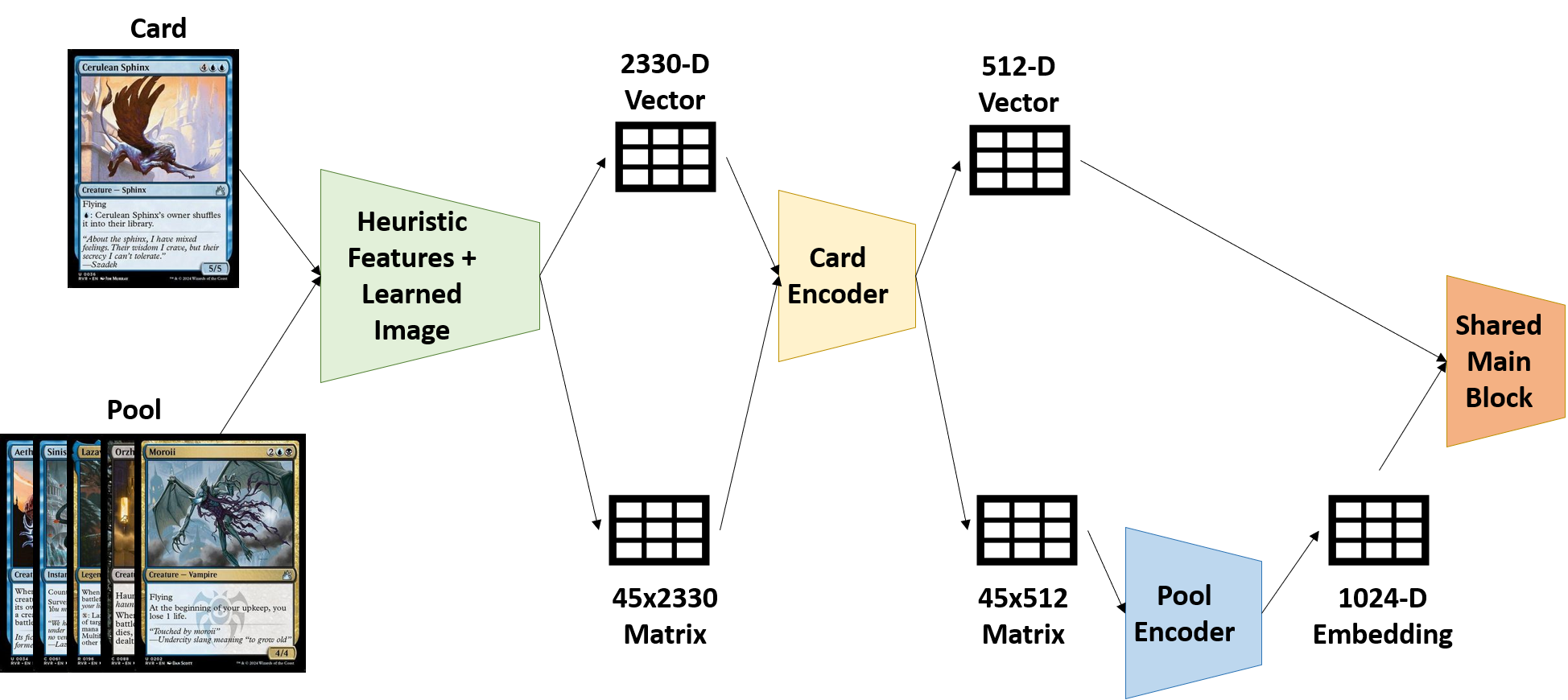}
    \caption{Weight-shared output layer (used in triplet loss).}
    \end{subfigure}%
    ~ 
    \begin{subfigure}[t]{0.5\textwidth}
        \centering
        \includegraphics[width=\textwidth]{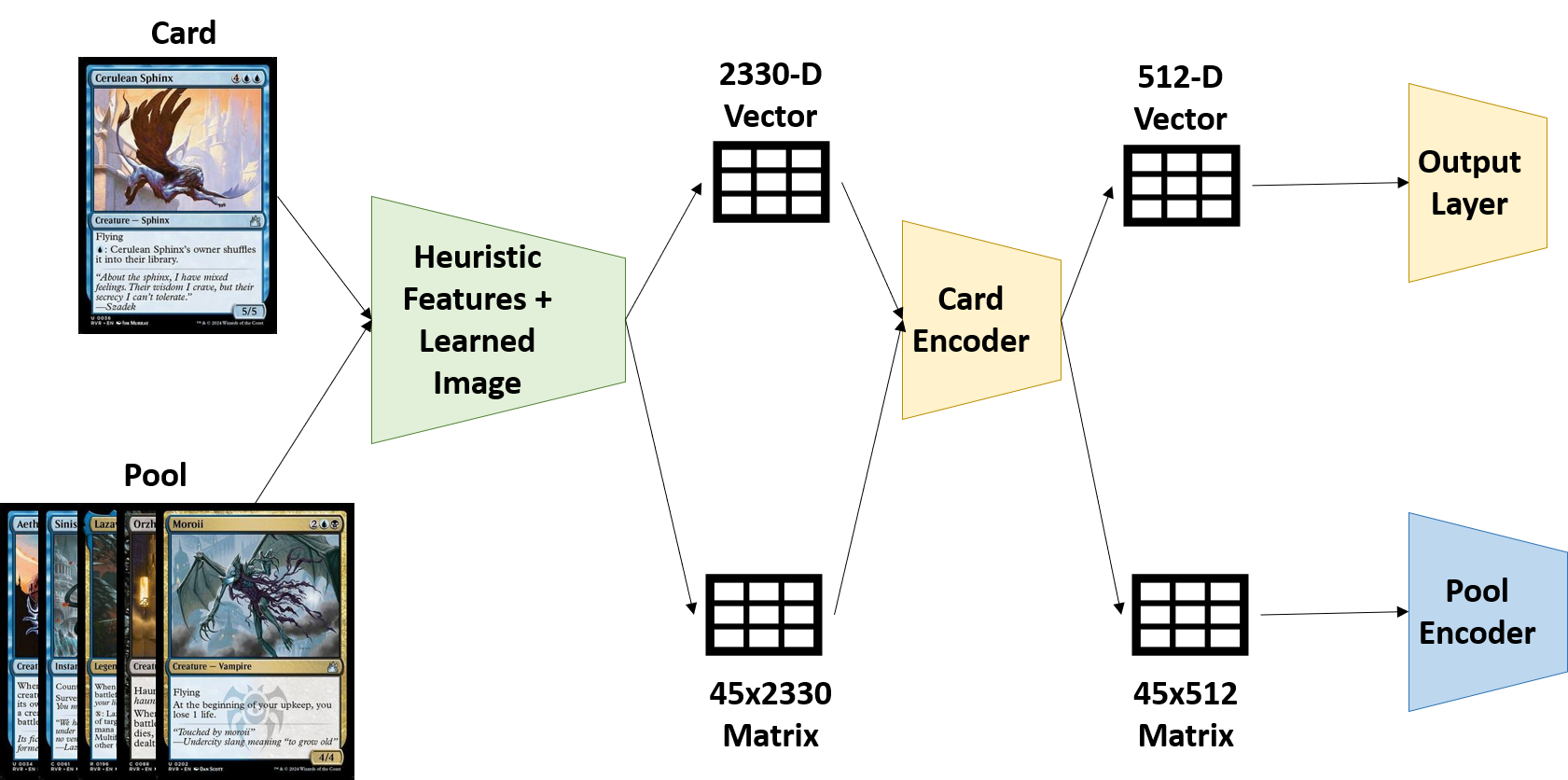}
    \caption{Separate output layers (used in InfoNCE loss).}
    \end{subfigure}
    \vspace{4mm}
    \caption{Differences between the forward pass pipeline in triplet loss-based networks and CLIP-based ones. When training with the triplet loss, we use a weight-shared output layer to accommodate a Siamese network structure. When using the InfoNCE loss, we reuse the card encoder when encoding a pool, but the main network blocks and final outputs are separate.}
    \label{fig:networks}
\end{figure*}

While it is not possible to entirely equalise the hyperparameters across settings, we aim to minimise the interference of parameter choice on the results. In addition, we realise that computation speed comparisons are hardware and implementation-specific. Still, we believe that our results provide useful insights into the problem setting. For many architectural choices, we modelled our experiments after the previous work by \citet{bertram2021predicting}.

\subsubsection{Card representation}
\textit{Magic: The Gathering} cards possess numerous attributes which are hard to properly represent to enable meaningful feature extraction. Crucially, cards consist of a variety of categorical and numerical attributes (e.g. cost, power, colour), but also contain important natural language text. To represent a single card, we encode all numerical and categorical features in a vector. We concatenate this with a BERT-generated \cite{devlin2018bert} embedding of the card text and an autoencoder-generated latent representation of the full card visual. Altogether, this results in a 2330-dimensional vector.

Card pools, i.e. a list of individual cards, are represented as matrices of 45 cards, as a pool can maximally be 45 elements long. Rows of the matrix that belong to yet-unchosen cards are set to 0. Sequence-based approaches, e.g. RNNs or transformers, might be suitable here but we simply process each pool matrix with convolutions.

\subsubsection{Neural Networks}

Single cards and card pools differ in their representation which necessitates different neural network structures to learn embeddings (see Figure~\ref{fig:networks}). We call those specialised networks \textit{card encoder} and \textit{pool encoder}, which ultimately serve to translate the two differently represented inputs to a shared embedding space. When using the triplet loss, the encoding networks are smaller due to the following \textit{shared main block}. For the InfoNCE variant, we do not use a \textit{Shared Main Block} and rather train two larger \textit{card encoders} and \textit{pool encoders}. While not strictly necessary, we experimentally found that using the \textit{card encoder} in the forward pass of pools aids training, as pools are simply lists of cards.

\textbf{Card encoders} are fully connected networks with 5 or 10 layers with 1024 neurons each, connected by normalisation and ELU-activations.
\textbf{Pool encoders} are convolutional neural networks consisting of 3 or 5 layers with 128 filters, which are connected in the same manner. For the triplet-based network, the main block consists of 5 additional fully connected layers, while for the InfoNCE network, the card output is a single layer. Both networks are trained with stochastic gradient descent on their respective loss function and a learning rate of 0.0003.

\subsection{Results}
\label{sec:results}

Finally, we experimentally compare the outlined approaches. All experiments are run on a single Nvidia A100 80GB GPU. We decided on the following different methods:

\begin{enumerate}
    \item Standard InfoNCE loss. This generates a cosine similarity matrix as seen in Figure~\ref{fig:overview}(a) and computes the cross entropy loss in both dimensions of the matrix, which are summed to generate the overall loss. 
    \item InfoNCE loss with a paired Sigmoid loss on the cosine similarities instead of using Softmax and the cross entropy loss \cite{zhai2023sigmoid}. This version alleviates some of the previously mentioned issues with the InfoNCE loss, e.g. multiple matching pairs, but still suffers from comparisons out of context.
    \item Our proposed adapted InfoNCE. As seen in Figure~\ref{fig:overview}(b), we generate a matrix of cosine similarities but mask the majority of similarities before computing the loss. We still use the cross entropy loss but only row-wise, as multiple positive pairs can occur in columns and column-wise associations are context agnostic.
    \item Triplet loss with random mining. For every training sample, we randomly select a negative example out of the available options.
    \item Triplet loss with hard mining. Before computing the loss, this method computes the distance of each negative example to the pool. The closest one, i.e. the one regarded as the best, is used for computing the loss, which provides additional difficulty.
    \item Triplet loss with all available triplets. Here, we generate all possible triplets for every sample. While this leads to a range of triplets with different difficulties, it bloats the computation time massively. To fit this approach into the GPU memory, we had to down-scale the batch size by a factor of 10. Thus, this setting should be regarded with caution.
\end{enumerate}

\begin{table}[h]
\caption{Comparison of the different approaches. As a performance measure, we use top-1 accuracy of predicting the correct choice on a held-out test set as well as training time per epoch. We see that the unchanged InfoNCE does not perform well in this task due to numerous inaccurate comparisons in the matrix, but our proposed adaptions lead to the best overall results. \textit{Triplet loss random mining} is fastest due to less computation overhead in the similarity matrix. \textit{all mining} should be regarded with caution as we had to downsize the batch size by a factor of 10, thus strongly influencing the results. Interestingly, \textit{hardest mining} leads to worse accuracy than \textit{random mining}}
\label{tab:results}
\resizebox{\columnwidth}{!}{
\begin{tabular}{ccccc}
\toprule
Method & Top-1 Accuracy & Training time per epoch\\
\toprule
Standard InfoNCE & 54.24\% & 52:47min\\
Sigmoid InfoNCE & 68.11\% & 53:30min\\
Adapted InfoNCE & \textbf{68.80\%} & 52:45min\\
Triplet loss random mining& 67.23\% & \textbf{47:31min}\\
Triplet loss hardest mining& 66.56\% &1:07:35h\\
Triplet loss all mining& 65.98\% & 11:51:24h\\
\bottomrule
\end{tabular}}
\end{table}

Table~\ref{tab:results} shows the results from our experiments. Here, we see a confirmation of our initial assumption -- the \textit{standard InfoNCE} loss without adaptions does not serve the task well. The matrix generated for it consists of numerous comparisons which are not based on the given data, thus greatly influencing the valid preferences. Using a \textit{sigmoid} loss leads to vastly better results, likely due to handling multiple positive pairs for one item, but still underperforms. From the \textit{triplet-based} methods, surprisingly \textit{random mining} performs best both in terms of training time and accuracy. While it obviously is the fastest due to adding no additional computation, it outperforming \textit{hard mining} is unexpected. Finally, our proposed \textit{adapted InfoNCE} reaches the highest accuracy of all tested methods. It introduces only irrelevant computation overhead compared to \textit{standard InfoNCE} and is only ~10\% slower than using single negative examples rather than a whole row of comparisons.

\section{Conclusion}

In this research, we aim to provide a novel way to frame contextual preferences by utilising the InfoNCE loss function. For numerous domains, preferences between items depend on specific situations, thus making it impossible to compare them in a vacuum. To alleviate this, we adapt the batch construction technique from CLIP \cite{radford2021learning} to only train on preferences explicitly contained in the training data.

We experimentally show (Section~\ref{sec:results}) that this technique outperforms both unaltered CLIP, CLIP with a \textit{sigmoid} loss function, and triplet loss-based models. The \textit{sigmoid-based} version is the second best in our experiments, as it also alleviates a subset of the problems of \textit{standard InfoNCE}, but our proposed method results in better performance. Our method adds no relevant computation time compared to the other InfoNCE methods and is thus strictly better-versed for the task at hand.

\textit{Triplet loss}-based methods had surprising differences in accuracy depending on the triplet mining technique. Random mining, i.e. \textit{randomly} choosing one of the available negative examples, is unsurprisingly fastest but also leads to the best accuracy. \textit{Hard mining} was assumed to lead to higher accuracy but required additional computation with a decreased accuracy. We speculate that a mix of both, e.g. first mining randomly and later in training switching to hard mining, might lead to higher accuracy, but such additional experiments are left for future work. Unsurprisingly, using \textit{all negative examples} in different triplets leads to poor results, as this greatly bloats training time due to reductions in batch size. In addition, this led to a stark performance degradation.

While only regarding a single domain here, collectable card games, our proposed adaptions translate to many domains. Preferences, or other metrics which be can be represented by an embedding space, often occur in specific contexts, which hinders the general capability of the InfoNCE loss. Altogether, this research forms a baseline for future work on contextual preferences, a highly relevant research question.

\clearpage
\section{Future Work}

While we show in this work that the InfoNCE loss can be adapted to the given problem with few changes, we speculate that overall more promising methods are possible. Masking invalid entries in the cosine similarity matrix works from a computational perspective but leads to a large number of unnecessary items. Instead, we aim to work towards an approach that explicitly adds the context of the preference to the loss computation while keeping the advantages of the InfoNCE loss.

\section*{Acknowledgements}
We want to thank Günter Klambauer for suggesting the use of the InfoNCE loss for this project and for his early feedback on our approach.

\bibliography{bib/mybibfile,bib/pl-book,bib/preferences,bib/jf}

\end{document}